\documentclass[journal]{IEEEtran}
\usepackage{amsmath,amsfonts,amssymb}
\usepackage{algorithmic}
\usepackage{array}
\usepackage[caption=false,font=scriptsize,labelfont=sf,textfont=sf]{subfig} 
\usepackage{textcomp}
\usepackage{stfloats}
\usepackage{url}
\usepackage{verbatim}
\usepackage{graphicx}
\usepackage{tabularx}
\usepackage{marvosym}
\usepackage[dvipsnames, table]{xcolor}
\usepackage{cite}
\usepackage[colorlinks=true, urlcolor=Black,linkcolor=blue, citecolor=blue]{hyperref} 
\usepackage{orcidlink}
\usepackage{braket}
\usepackage{booktabs}
\usepackage{threeparttable}
\usepackage{multirow} 
\usepackage{makecell}
\usepackage{pifont}
\usepackage[normalem]{ulem}
\definecolor{darkgreen}{RGB}{0,128,0}  
\newcommand{\hhline}{\noalign{\vskip 1pt}\hline\noalign{\vskip 1pt}}

\def\BibTeX{{\rm B\kern-.05em{\sc i\kern-.025em b}\kern-.08em
    T\kern-.1667em\lower.7ex\hbox{E}\kern-.125emX}}
\usepackage{balance}

\begin{document}
\title{Minutiae-Anchored Local Dense Representation for Fingerprint Matching}

%

\author{Zhiyu Pan$^{\orcidlink{0009-0000-6721-4482}}$,
  Xiongjun Guan$^{\orcidlink{0000-0001-8887-3735}}$, 
  Yongjie Duan$^{\orcidlink{0000-0003-3741-9596}}$,
	Jianjiang Feng$^{\orcidlink{0000-0003-4971-6707}}$, ~\IEEEmembership{Member, IEEE}, 
	and Jie Zhou$^{\orcidlink{0000-0001-7701-234X}}$, ~\IEEEmembership{Fellow, IEEE}
  \thanks{
    This work was supported in part by the National Natural Science Foundation of China under Grant 62376132 and 62321005. (\emph{Corresponding author: Jianjiang Feng}.)}
  \IEEEcompsocitemizethanks{
  \IEEEcompsocthanksitem
  The authors are with Department of Automation, Tsinghua University, Beijing 100084, China (e-mail: \url{pzy20@mails.tsinghua.edu.cn}; \url{gxj21@mails.tsinghua.edu.cn};
  \url{duanyj13@tsinghua.org.cn};  \url{jfeng@tsinghua.edu.cn}; \url{jzhou@tsinghua.edu.cn}).}
}

%
%

\markboth{Journal of \LaTeX\ Class Files,~Vol.~14, No.~8, June~2025}%
{Preprint of the \textit{DMD} Pro.}
%



\maketitle

\begin{abstract}
Fingerprint matching under diverse capture conditions remains a fundamental challenge in biometric recognition. To achieve robust and accurate performance in such scenarios, we propose DMD, a minutiae-anchored local dense representation which captures both fine-grained ridge textures and discriminative minutiae features in a spatially structured manner. Specifically, descriptors are extracted from local patches centered and oriented on each detected minutia, forming a three-dimensional tensor, where two dimensions represent spatial locations on the fingerprint plane and the third encodes semantic features. This representation explicitly captures abstract features of local image patches, enabling a multi-level, fine-grained description that aggregates information from multiple minutiae and their surrounding ridge structures. Furthermore, thanks to its strong spatial correspondence with the patch image, DMD allows for the use of foreground segmentation masks to identify valid descriptor regions. During matching, comparisons are then restricted to overlapping foreground areas, improving efficiency and robustness. Extensive experiments on rolled, plain, parital, contactless, and latent fingerprint datasets demonstrate the effectiveness and generalizability of the proposed method. It achieves state-of-the-art accuracy across multiple benchmarks while maintaining high computational efficiency, showing strong potential for large-scale fingerprint recognition. Corresponding code is available at \href{https://github.com/Yu-Yy/DMD}{https://github.com/Yu-Yy/DMD}.
\end{abstract}


%
\IEEEpeerreviewmaketitle

\section{Introduction}
%
%
%

\IEEEPARstart{M}{inutiae} are distinctive local features in fingerprints, typically categorized as ridge endings and bifurcations, marking the abrupt termination or splitting of ridge flows. As the most widely used features in both manual and automated fingerprint recognition, minutiae provide concise and discriminative cues that remain highly stable across different impressions of the same finger \cite{maltoni2022handbook}. Their definition aligns with features utilized by human experts, ensuring compatibility across various algorithms. The distribution of minutiae is highly random, providing sufficient discriminative information for identification, and enabling good recognition performance even with traditional point-matching algorithms. Due to these advantages, minutiae-based fingerprint recognition found practical applications early in the development of artificial intelligence (around the 1970s) \cite{cole2004history}. To date, minutiae-based methods remain a popular approach for fingerprint matching. 

In the early days, fingerprint matching based on minutiae treated the fingerprint as a collection of minutiae, where the geometric similarity between two minutiae sets was considered the fingerprint's similarity score. For example, Delaunay triangulation \cite{munoz2013fingerprint} or quadrilateral tessellation \cite{iloanusi2011indexing} could be used to construct structural relationships between adjacent minutiae. To better capture local features, researchers also explored using local feature representations, anchoring the encoding of minutiae neighborhood relationships around each minutia \cite{cappelli2010minutia}. However, matching solely based on the relationships between minutiae requires a sufficient number of true minutiae. In cases of poor-quality or incomplete fingerprints (e.g., latent fingerprints), where minutiae are often misidentified or missed, the performance of such matching methods drops significantly.


To enhance the robustness of minutiae-based fingerprint matching, researchers have explored incorporating ridge-based texture features into the representation. For instance, orientation fields have been used as local descriptors \cite{Tico2003fingerprint, feng2008combining}, and ridge textures from patches centered and aligned on minutiae have served as descriptors for the corresponding anchor points \cite{zhou2013fingerprint}. These manually designed representations have shown decent matching performance by capturing complementary local cues. However, due to their heuristic nature, they often struggle to generalize well to challenging cases such as latent fingerprints, where image quality is low and distortion is common. 

With the advancement of deep learning and its superior capacity for robust feature extraction, recent studies have focused on learning minutiae descriptors directly from fingerprint images. In particular, a number of methods \cite{cao2018texture, cao2019automated, cao2020end, MinNet, grosz2023latent} have been proposed for latent fingerprint matching, a particularly challenging scenario that demands high recognition accuracy. These approaches leverage deep networks to extract discriminative representations and have demonstrated promising performance. Building on this line of work, we revisit the design of minutiae descriptors and identify opportunities for enhancing existing feature representations. Most existing methods encode minutiae-centered patches into one-dimensional feature vectors \cite{cao2019automated, cao2020end, MinNet, grosz2023latent}. While compact, such representations overlook spatial information and lack a fine-grained correspondence to the fingerprint image. Consequently, they are less effective at suppressing background interference and are more vulnerable to ridge discontinuities and noise.


\begin{figure}[!t]
  \centering
  \includegraphics[width=.92\linewidth]{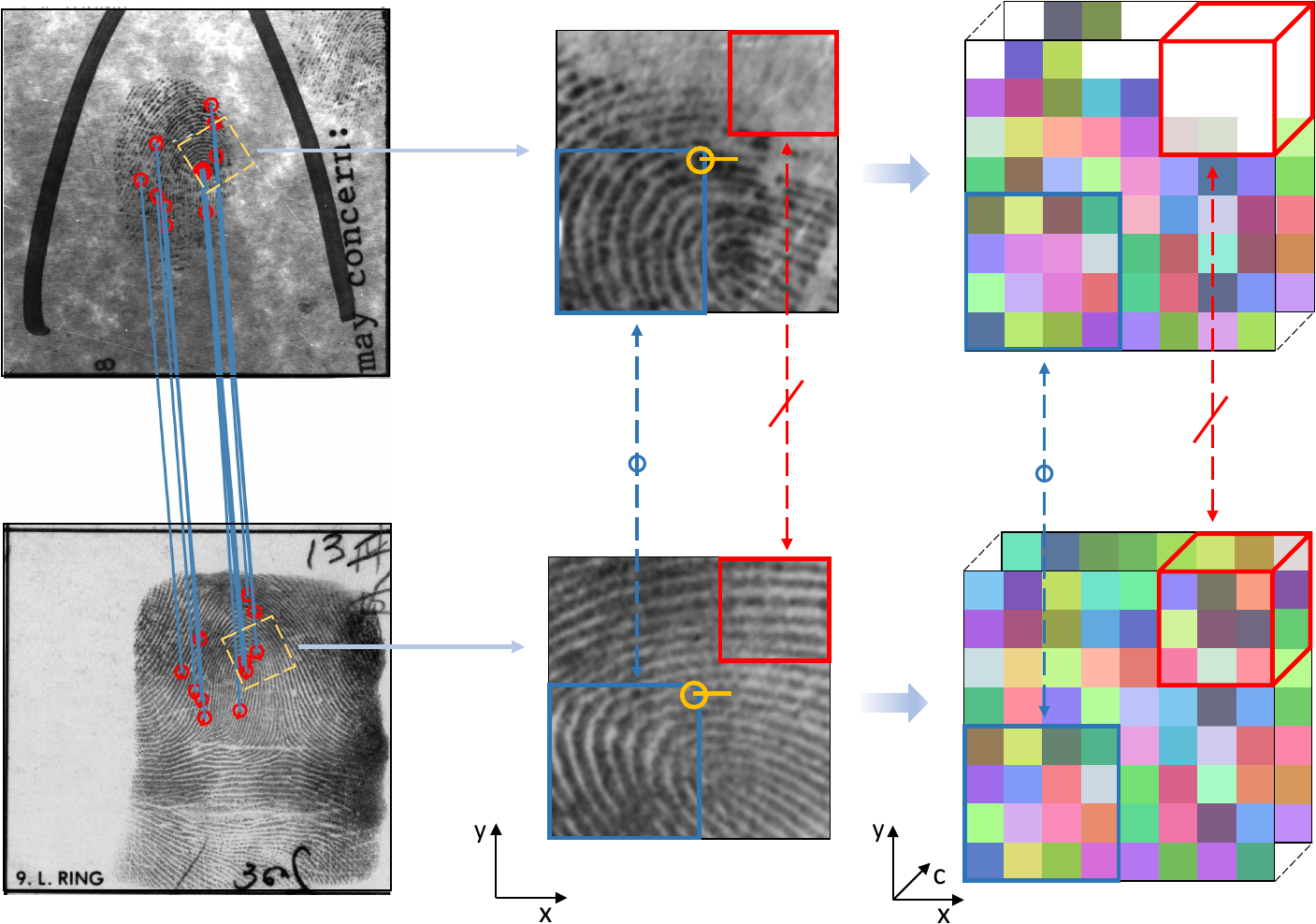}
  \caption{Illustration of the Dense Minutiae Descriptor (DMD). DMD is a local dense representation with spatial structure, where the descriptor space is aligned with the original fingerprint image, enabling effective suppression of background regions. ``\includegraphics[height=0.015\linewidth]{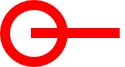}'' denotes detected minutiae, and connecting lines indicate matched pairs. ``\includegraphics[height=0.02\linewidth]{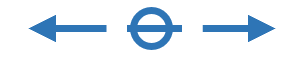}'' indicates that the rectangular regions in the two patches correspond to each other, while ``\includegraphics[height=0.02\linewidth]{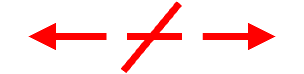}'' indicates that the rectangular regions do not correspond.} \label{fig:DMD_illustration}
\end{figure}

Therefore, our work focuses on innovative feature representation, enhancing local feature representation capabilities and robustness. A common approach for extracting minutiae representations involves translating, rotating, and cropping a local image patch around each minutia, from which the corresponding descriptor is derived. To improve representational capacity and robustness, we adopt a descriptor format similar to the cylinder encoding used in \cite{cappelli2010minutia}, but with features learned through deep learning, resulting in a fine-grained representation of fingerprint regions. Thus, we propose the Dense Minutia Descriptor (DMD), which provides a minutiae-anchored local dense representation. 

The proposed dense representation is a three-dimensional tensor, where two dimensions preserve the spatial layout and the third encodes abstract semantic features (Fig.~\ref{fig:DMD_illustration}). Since all patches are spatially normalized based on minutiae position and orientation, the resulting DMD descriptors are spatially aligned, making the matching process invariant to global transformations while maintaining strong local spatial-texture associations for enhanced discriminability. To further exploit this spatial structure, we incorporate sinusoidal positional embeddings to boost the distinctiveness of each descriptor location. Besides, we adopt a dual-stream architecture that extracts features from both ridge texture and neighboring minutiae, inspired by \cite{feng2008combining, DeepPrint}, and concatenate them to construct the final DMD. Additionally, a segmentation mask is predicted to explicitly indicate valid foreground regions and suppress background noise, as illustrated in the first row of Fig.~\ref{fig:DMD_illustration}. It can be observed that non-fingerprint areas are also marked as invalid in the descriptor space. During matching, only features within the overlapping foreground regions are considered, effectively reducing the impact of ridge discontinuities.

Extensive experiments on rolled, plain, contactless, and latent fingerprints demonstrate that our method surpasses existing minutiae-based representations \cite{cappelli2010minutia,nist2020verifinger,MinNet,cao2019automated} in both accuracy and efficiency. Ablation studies further validate the effectiveness of the DMD design. This work extends our previous conference paper \cite{pan2024latent}, which focused exclusively on latent fingerprints, by introducing several methodological enhancements—such as incorporating an input distribution normalization module into the network and adopting more diverse data augmentation strategies during training. In addition, we conduct more comprehensive evaluations across a wider range of fingerprint modalities and perform in-depth analyses of DMD's effectiveness. The updated version achieves higher accuracy, and the relaxation-based refinement module has been reimplemented for significantly improved matching efficiency. In summary, the main contributions of this work are as follows:
\begin{itemize}
  \item We propose DMD, a dense, variable-length fingerprint representation centered on minutiae, which preserves spatial structure and enhances local discriminative power.
  \item We introduce a series of design strategies for local representation—including dual-stream structure, positional encoding, and foreground masking—to improve robustness against background noise, ridge discontinuities, and partial overlaps.
  \item Extensive experiments on multiple fingerprint modalities demonstrate that our method outperforms existing minutiae-based approaches in both accuracy and efficiency, and shed light on the effectiveness of its design.
\end{itemize}

\section{Related Literature}
This section reviews deep learning-based fingerprint matching methods, which are broadly categorized into global-level and local-level approaches.
\vspace{-0.2cm}
\subsection{Global-level}
Global-level fingerprint matching methods aim to extract holistic descriptors from entire fingerprint images. Approaches like DeepPrint~\cite{DeepPrint} integrated a Spatial Transformer Network (STN) for implicit alignment, while AFR-Net~\cite{grosz2024afrnet} incorporated ViT-based encoders to introduce global attention. More recently, Gu et al.\cite{gu2022latent} and Pan et al.\cite{FDD} utilized fingerprint pose estimation to normalize inputs and extract fixed-length dense representations that are moderately localized and robust to distortions.

However, the effectiveness of global representations heavily depends on alignment accuracy, which is often unreliable for latent or partial fingerprints. To bypass this, several verification-oriented methods~\cite{PFVNet, guan2025joint, ifvit} directly compare fingerprint pairs via registration or cross-attention mechanisms. While these approaches improve accuracy in verification settings, they are computationally expensive and impractical for large-scale identification due to low matching efficiency.

\vspace{-0.2cm}
\subsection{Local-level}
Local-level fingerprint matching methods rely on comparing local patches represented by their corresponding descriptors. To comprehensively capture local fingerprint details, Cao et al. \cite{cao2019automated} used both densely sampled ridge points and minutiae as anchors to extract multiple templates from various enhanced images. While this improves matching robustness, it introduces substantial storage and computational overhead. Their subsequent work \cite{cao2020end} reduced the number of templates and applied quantization to compress their size, but the computational cost remained high. Similarly, Gu et al. \cite{gu2021latent} uniformly sampled anchor points across the fingerprint and performed patch-level pairwise comparisons, further increasing matching complexity. To improve the storage and matching efficiency of variable-length representations, {\"O}zt{\"u}rk et al. \cite{MinNet} proposed using only minutiae as anchors and introduced a local descriptor extraction framework. To enhance representation power, they predicted a minutiae map within each patch as an auxiliary task—a strategy that echoes the minutiae structure modeling in Cappelli et al \cite{cappelli2010minutia}. All these methods rely on one-dimensional feature vectors as compact representations. However, achieving high matching accuracy often demands a large number of anchor points or complex refinement procedures.

Our proposed DMD also adopts minutiae as anchors, but fundamentally differs from previous designs by introducing a novel spatially structured dense representation. Unlike traditional one-dimensional vectorized descriptors, DMD preserves the spatial layout of local features within each minutia-centered patch, enabling richer spatial expressiveness and multi-level fine-grained positional representation. To further enhance its descriptive power, we adopt a dual-stream architecture that integrates complementary cues from ridge texture and minutiae context. The matching procedure is also optimized for efficiency. As a result, DMD combines strong representational capacity with low computational cost and high matching efficiency, making it both storage- and computation-friendly.

\begin{figure*}[!t]
  \centering
  \includegraphics[width=0.924\linewidth]{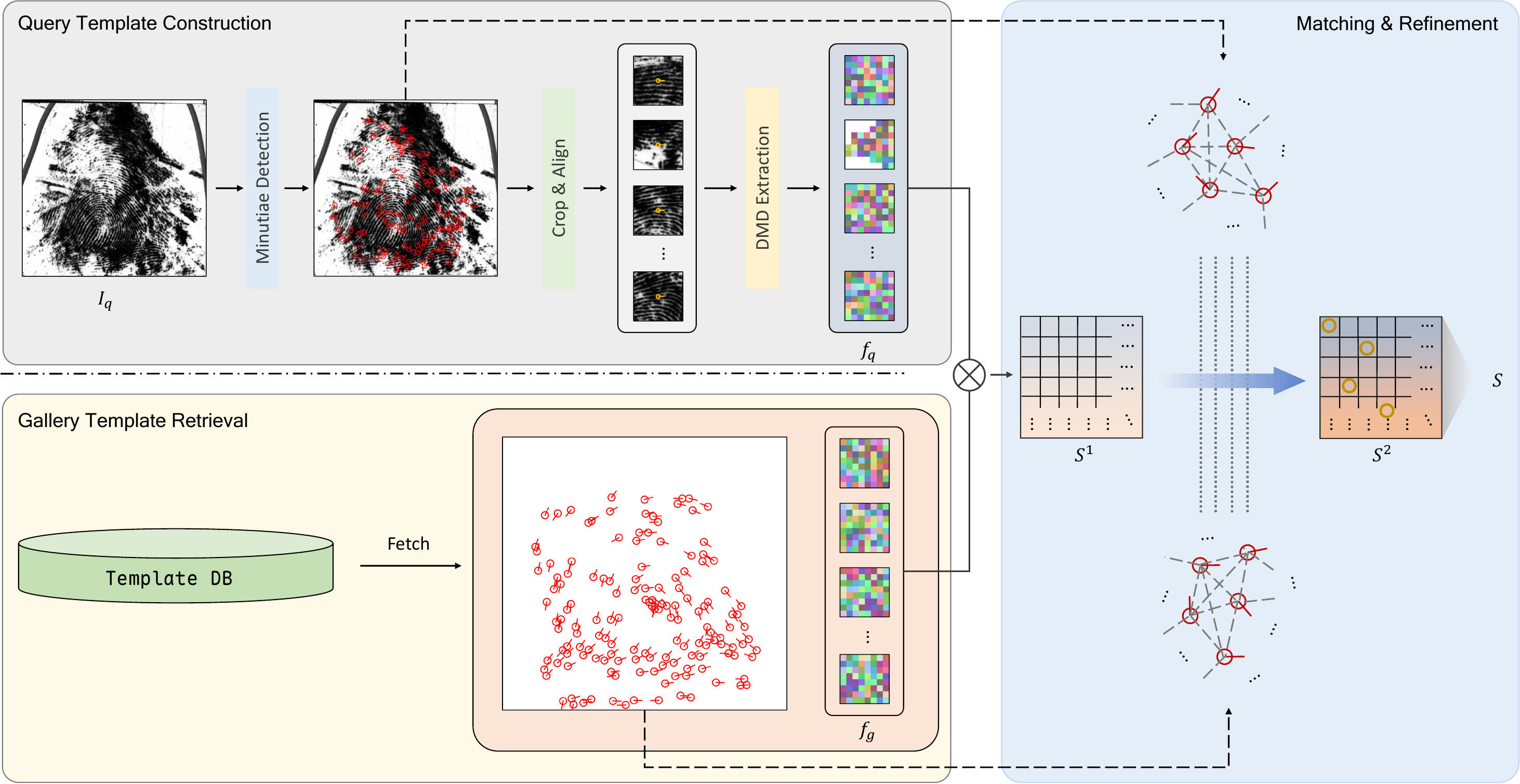}
  \caption{DMD matching pipeline. Each fingerprint image is divided into minutiae-centered and aligned patches, yielding a variable number of DMDs. The extracted descriptors are compared with stored gallery templates, followed by relaxation refinement for the final score.}\label{fig:DMD_Matching}
\end{figure*}

\section{Method}

\subsection{Overview}
The overall DMD matching pipeline is illustrated in Fig. \ref{fig:DMD_Matching}. Given a pair of fingerprint images, minutiae are first extracted using existing methods such as \cite{nist2020verifinger, FDD}. For each minutia, a local patch is cropped by centering on its location and aligning its orientation to the right. These patches are then passed through the descriptor extraction network to generate dense representations and corresponding segmentation masks, forming the DMDs. Matching is performed between the DMDs and the associated minutiae positions and orientations from the query and gallery. Initial similarity scores are computed by comparing representations, and subsequently refined via a relaxation-based process that enforces geometric consistency among minutiae. The final fingerprint matching score is obtained after this refinement. The following sections detail the descriptor extraction network, matching algorithm, and other implementation specifics.

\subsection{Local Dense Representation Extraction}
\begin{figure*}
  \centering
  \includegraphics[width=.92\linewidth]{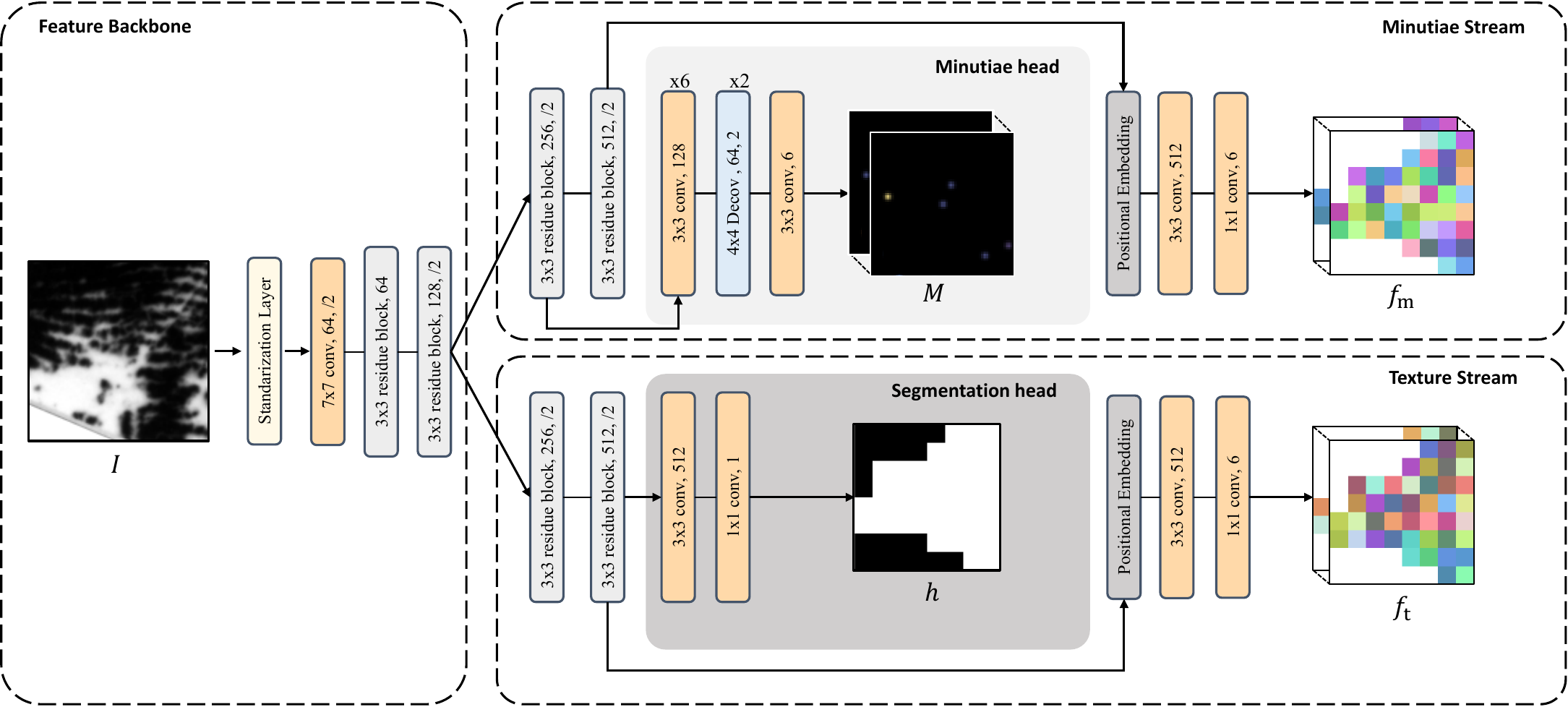}
  \caption{A schematic of the local representation extraction network for DMD generation. The content boxes include the operation names, number of output channels, and spatial scales (omitted if equal to 1).} 
  \label{fig:DMD_extraction}
\end{figure*}

We propose a dual-stream extraction network built upon a modified ResNet-34 \cite{resnet} to generate expressive local dense representations under multi-task supervision. Each stream is guided by an auxiliary task—fingerprint region segmentation and minutiae map prediction, respectively—to enhance the extraction of texture-related and minutiae-related features. To further improve the discriminative power at each spatial location, we incorporate 2D sinusoidal positional embeddings \cite{vaswani2017attention} prior to descriptor encoding. The input fingerprint patch $I$ has a size of $128 \times 128$ pixels at 500 ppi resolution, and the resulting DMD is of shape $\mathbb{R}^{2C\times8\times8}$, with each position representing approximately a $16 \times 16$ pixel region in the original fingerprint, and $C$ denotes the channel dimension of each individual stream representation. The specific architecture of the extraction network is illustrated in Fig. \ref{fig:DMD_extraction}.

\subsubsection{Feature Backbone}
The basic feature backbone for extracting initial abstract representations consists of the stem module of ResNet-34 and the first two residual blocks. To preserve fine-grained fingerprint ridge details, the max pooling layer following the initial convolution is removed. This modification results in a feature map with a spatial resolution of $\frac{1}{4}$ of the input image size, which serves as the shared input for the two subsequent branches. Additionally, a standardization layer is introduced before the backbone to normalize the input images to zero mean and unit variance, enhancing training stability and ensuring consistent contrast levels across input samples.

\subsubsection{Texture Stream}
In the Texture Stream, the final two residual blocks of ResNet-34 further process the shared features, producing a feature map with a spatial resolution of $\frac{1}{16}$ of the input image size, which matches the resolution of the final dense representation. To guide the learning of texture-aware features, 2D sinusoidal positional embeddings are added to this feature map before predicting the texture-related dense representation $f_\text{t} \in \mathbb{R}^{C \times 8 \times 8}$ through a dedicated lightweight convolutional head. The fingerprint segmentation mask in descriptor space $h \in \mathbb{R}^{1 \times 8 \times 8}$ is predicted in parallel using a separate head applied directly to the same feature map.

\subsubsection{Minutiae Stream}
In the Minutiae Stream, the shared features are first further processed by the final two residual blocks of ResNet-34. Unlike the Texture Stream, the minutiae map prediction branch does not originate from the final output; instead, it branches off earlier and passes through a series of convolutional and deconvolutional layers to progressively increase the spatial resolution. This produces a minutiae map $M\footnote{The minutiae map follows the same configuration as in \cite{DeepPrint}, including the definition of channels and the representation format.} \in \mathbb{R}^{6 \times 64 \times 64}$, which corresponds to a resolution of one-half of the input image size and provides coarse localization and angles of fingerprint minutiae. Meanwhile, the remaining features continue through the main branch, where 2D sinusoidal positional embeddings are added before predicting the minutiae-related dense representation $f_\text{m} \in \mathbb{R}^{C \times 8 \times 8}$ using a dedicated lightweight convolutional head. 

Finally, the DMD $f \in \mathbb{R}^{2C \times 8 \times 8}$ is obtained by concatenating the texture-related descriptor $f_\text{t}$ and the minutiae-related descriptor $f_\text{m}$, followed by element-wise modulation using the segmentation mask $h$:
\begin{equation}
  \label{eq:dmd}
  f = (f_\text{t} \mathbin\Vert f_\text{m}) \odot h \;,
\end{equation}
where $\mathbin\Vert$ denotes channel-wise concatenation and $\odot$ represents the Hadamard product.

\vspace{-0.1cm}
\subsection{Training Objective} 
\subsubsection{Descriptor Loss} \label{sec:desc_loss}
Considering that our proposed DMD adopts a dense representation format, it is essential to ensure two key properties: (1) the ability to distinguish between patches originating from different minutiae, and (2) the consistency of features within valid regions across genuine matching samples. To address the first property, which is fundamental for discriminative learning, we employ CosFace loss~\cite{wang2018cosface} to supervise the texture stream $f_\text{t}$ and the minutiae stream $f_\text{m}$ separately, ensuring that the descriptors can effectively differentiate between patches from distinct minutiae. The classification loss is defined as $\mathcal{L}_\text{cls} = \mathcal{L}_\text{cls}^\text{m} + \mathcal{L}_\text{cls}^\text{t}$. Each dense descriptor is flattened and passed through a fully connected layer to classify it into one of $K$ minutiae-centered patch classes, where $K$ corresponds to the number of distinct minutiae identities defined across the training set. To enforce the second property, feature consistency under intra-class variations, we generate paired original and corresponding masked patches using segmentation maps from the DPF dataset~\cite{duan2023estimating}. These two patches are then further augmented using different random transformations, as described in Sec.~\ref{sec:imple}. A consistency loss is applied to encourage similarity within the overlapping valid regions:
\begin{equation}
  \label{eq:cons}
  \mathcal{L}_\text{cons} = \frac{1}{|h_\text{valid}|} \sum\nolimits_{(i,j)\in h_\text{valid}} \left\| f_\text{ori}^{ij} - f_\text{mask}^{ij} \right\|_2^2 \;,
\end{equation}
where $f_\text{ori}$ and $f_\text{mask}$ denote descriptors from the original and masked patches, respectively, and $h_\text{valid} = h_{\text{ori} \cap \text{mask}}$ denotes the set of overlapping valid positions defined by the corresponding foreground masks.

\subsubsection{Auxiliary Loss}
For the auxiliary tasks, the network is supervised to predict both the foreground mask and the minutiae map. We use a binary cross-entropy loss $\mathcal{L}_\text{mask}$ for foreground segmentation and a mean squared error loss $\mathcal{L}_\text{mnt}$ for minutiae map regression. The ground-truth annotations for both tasks are obtained using VeriFinger v12.0 \cite{nist2020verifinger}.

Therefore, the overall loss for extracting the local dense representation DMD is defined as:
\begin{equation}
  \mathcal{L} = \mathcal{L}_\text{cls} + \lambda_\text{cons}\mathcal{L}_\text{cons} + \lambda_\text{mask}\mathcal{L}_\text{mask} + \lambda_\text{mnt} \mathcal{L}_\text{mnt}\;,
  \label{eq:loss}
\end{equation}
where $\lambda_\text{cons}$, $\lambda_\text{mnt}$, and $\lambda_\text{mask}$ are corresponding weights to balance the loss components.

\subsection{Fingerprint Matching}
For a pair of fingerprint images $(I_q, I_g)$, we extract the corresponding DMDs  $\{f_q^i | i\in[0,r)\}$ and $\{f_g^j | j\in[0,p)\}$ from minutiae-aligned cropped patches using Eq.~\ref{eq:dmd}, where $r$ and $p$ denote the number of minutiae detected in $I_q$ and $I_g$, respectively. A similarity score matrix $S^1_{(q,g)}$ is then computed between the two sets of local dense descriptors. Each entry in the matrix represents the similarity between a pair of descriptors, calculated using cosine similarity over their flattened features within the overlapping valid region. 
The similarity score between descriptors $f_q^i$ and $f_g^j$ is defined as:
\begin{equation}
S^{1}_{(q,g)}(i,j) = \frac{ \left\langle \left( f_q^i \odot h_g^j \right)^{\flat}, \left( f_g^j \odot h_q^i \right)^{\flat} \right\rangle }{ \left\| f_q^i \odot h_g^j \right\|_F \cdot \left\| f_g^j \odot h_q^i \right\|_F}\;, 
\label{eq:dmd_sim}
\end{equation}
where $h_q^i$ and $h_g^j$ are corresponding foreground masks; $\| \cdot \|_F$ is the Frobenius norm; and $(\cdot)^{\flat}$ denotes the flattening operation. For the binary version, where both the descriptor $f$ and mask $h$ are binarized, the similarity is computed using a natural and intuitive Hamming-distance-based formulation:
\begin{equation}
  S^{1}_{(q,g)}(i,j) = 1 - \frac{ \left\| (f_q^i\oplus f_g^j) \cap h_q^i \cap h_g^j\right\| }{\left\|h_q^i \cap h_g^j\right\|} \;.
  \label{eq:dmd_b_sim}
\end{equation} 

To incorporate the geometric consistency of matched minutiae, we refine the initial similarity matrix $S^1_{(q,g)}$ by adjusting each entry based on the spatial and directional relationships between the corresponding minutiae. Specifically, for each score $S^1_{(q,g)}(i, j)$, the refinement considers the geometric compatibility of minutiae $m_q^i$ and $m_g^j$, and propagates this compatibility to their neighboring pairs, resulting in an updated score matrix $S^2_{(q,g)}$. This refinement follows the Local Similarity Assignment with Relaxation (LSA-R) scheme dedicated in ~\cite{cappelli2010minutia}, which integrates the Hungarian algorithm with iterative updates guided by minutiae geometric relationships. We then select the top-$n_m$ matched minutiae pairs $\mathcal{M}$ based on their geometric consistency and compute the final matching score from the refined similarity matrix $S^2_{(q,g)}$. The number $n_m$ is adaptively determined by the number of detected minutiae in the query and gallery as:
\begin{equation}
  \label{eq:n_Set}
  n_m = n_\text{min} + \left\lfloor \frac{n_\text{max} - n_\text{min}}{1 + \exp\left(-\tau (\min(r, p) - \mu)\right)} \right\rceil,
\end{equation}
where $r$ and $p$ are the numbers of minutiae in the query and gallery fingerprints, respectively. The final fingerprint matching score $S(q, g)$ is computed by averaging the top $n_m$ scores from $S^2_{(q,g)}$:
\begin{equation}
  \label{eq:final_score}
  S(q, g) = \frac{1}{n_m} \sum\nolimits_{(i,j) \in \mathcal{M}} S^2_{(q,g)}(i,j).
\end{equation}

\subsection{Training Samples Preparation} 
To ensure effective training, we adopt a set of strategies to enforce accurate patch-level supervision. This includes establishing reliable minutiae correspondences and selecting patch pairs with minimal spatial redundancy within each fingerprint. Specifically, we use VeriFinger v12.0 \cite{nist2020verifinger} to extract minutiae and fingerprint segmentation maps, filtering out minutiae located in invalid regions. Initial minutiae correspondences between genuine pairs are obtained using the MCC descriptor \cite{cappelli2010minutia}. To further eliminate incorrectly matched pairs, we apply the RANSAC algorithm based on a 2D affine transformation estimated between the minutiae sets. To promote broader spatial coverage across the fingerprint area, we apply Farthest Point Sampling (FPS) to select a subset of up to five minutiae that are well-separated spatially. For each selected minutia, a local patch is extracted by translating and rotating the fingerprint image to align the minutia to the center and orient it horizontally to the right. We use a patch size of $128 \times 128$ pixels at 500 ppi resolution. 

\subsection{Implementation Details} \label{sec:imple}
We apply several data augmentation techniques to improve training and prevent overfitting. Random elastic distortions are generated using the model from Si et al. \cite{si2015detection}, and random histogram matching adjusts patch intensities to match distributions from the FVC2004 DB1A dataset. Rigid transformations include translations up to 10 pixels and rotations within $[-5^\circ, 5^\circ]$. We also add Gaussian noise (mean 0, std 5), Gaussian blur, wet/dry fingerprint simulation, and random line occlusion as in \cite{FLARE}. Each dense representation branch is set to $C=6$, resulting in a 12-channel DMD. The loss weights in Eq.~\eqref{eq:loss} are set to $\lambda_{\text{cons}}=0.00125$, $\lambda_{\text{mask}}=1$, and $\lambda_{\text{mnt}}=0.1$. We use AdamW with a learning rate of $3.5 \times 10^{-4}$, batch size 96, and apply L2 regularization. Training is performed on a single NVIDIA RTX 3090. In our experiments, the hyperparameters for relaxation-based matching are set to $n_\text{min} = 4$, $n_\text{max} = 12$, $\tau = 0.4$, and $\mu = 20$ when using minutiae extracted by VeriFinger \cite{nist2020verifinger}; for minutiae extracted by FDD \cite{FDD}, we use $n_\text{min} = 6$, $n_\text{max} = 14$, $\tau = 0.3$, and $\mu = 20$.

\begin{table*}[!t]
  \begin{threeparttable}
    \centering
    \caption{Fingerprint datasets used in our work.}
    \label{tab:datasets}
    \begin{tabular}{llllccc}
      \toprule
      \textbf{Type} & \textbf{Dataset} & \textbf{Sensor} & \textbf{Description} & \textbf{Usage} & \textbf{Genuine Pairs} & \textbf{Imposter Pairs} \\
      \midrule
      \multirow{2}{*}{Rolled} & NIST SD14 & Inking & 27,000 rolled pairs & train & - & -\\
      & NIST SD4 & Inking & 2,000 rolled pairs & test & 2,000 & 3,998,000 \\
      \hhline
      \multirow{2}{*}{Plain} & FVC2004 DB1A\tnote{a}  & Optical & 100 fingers, 8 impressions each & test & 2,800 & 4,950 \\
      & N2N Plain & Optical & 2,000 plain-rolled pairs & test & 2,000 & 3,998,000 \\
      \hhline
      \multirow{2}{*}{Partial} & FVC2002 DB3A\tnote{a} & Capacitive & 100 fingers, 8 impressions each & test & 2,800 & 4,950 \\
      & FVC2006 DB1A\tnote{b}  & Electric field  & 140 fingers, 12 impressions each & test & 9,240 & 9,730 \\
      \hhline
      \multirow{3}{*}{Latent} & NIST SD27\tnote{c} & - & 258 latent-rolled pairs (crime scene) & test & 258 & 2,764,470 \\
      & N2N Latent & - & 3,383 latent / 2,000 rolled (lab) & test & 3,383 & 6,762,617\\ 
      & THU Latent10K & - & 10,458 latent-plain/rolled pairs (crime scene) & test & 10,458 & 109,359,306 \\
      \hhline
      Contactless & PolyU CL2CB\tnote{d} & Optical/Camera & 336 fingers, 6 contact-based \& 6 contactless each & test & 12,096 & 4,052,160 \\
      \bottomrule
    \end{tabular}
    \begin{tablenotes}
      \item[a] Genuine pairs: $100 \times \binom{8}{2} = 2{,}800$; imposter pairs: $\binom{100}{2} = 4{,}950$.\vspace{0.1cm}
      \item[b] Genuine pairs: $140 \times \binom{12}{2} = 9{,}240$; imposter pairs: $\binom{140}{2} = 9{,}730$.\vspace{0.1cm}
      \item[c] The gallery set of THU Latent10K (10,458 plain or rolled fingerprints) is merged with the original gallery.\vspace{0.1cm}
      \item[d] Genuine pairs: $336 \times 6 \times 6 = 12{,}096$; imposter pairs: $336 \times 335 \times 6 \times 6 = 4{,}052{,}160$.
    \end{tablenotes}
  \end{threeparttable}
\end{table*}

\begin{figure*}[!t]
  \centering
  \subfloat[]{\includegraphics[height=.09\linewidth]{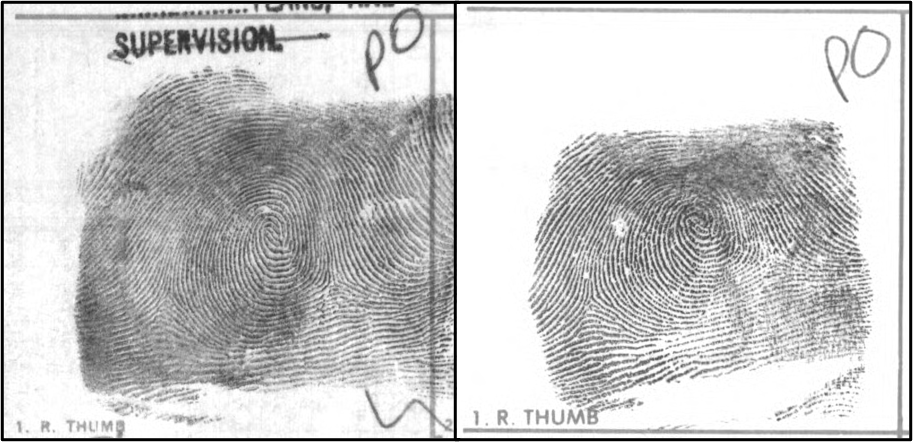}} \hfil
  \subfloat[]{\includegraphics[height=.09\linewidth]{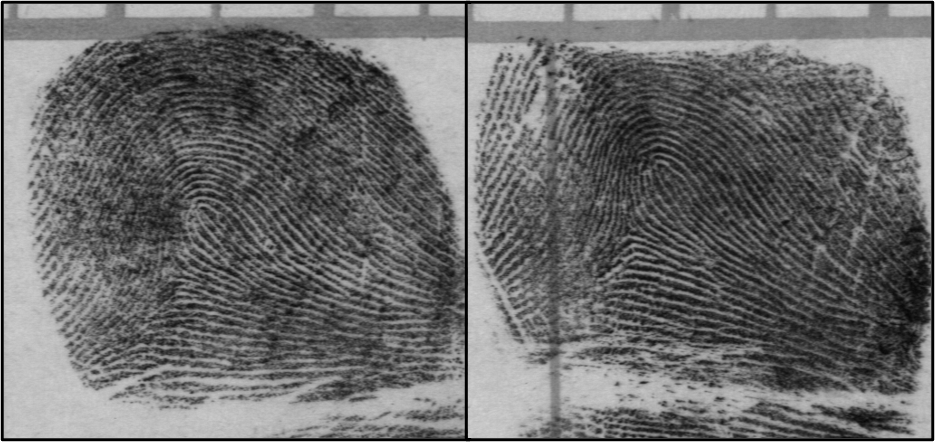}} \hfil
  \subfloat[]{\includegraphics[height=.09\linewidth]{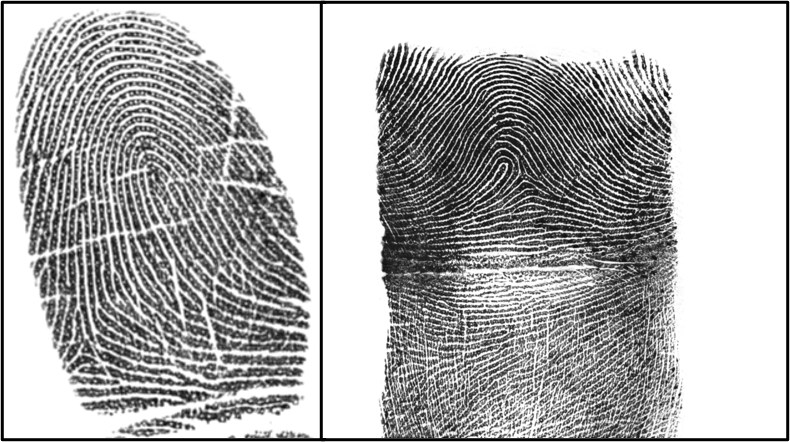}} \hfil
  \subfloat[]{\includegraphics[height=.09\linewidth]{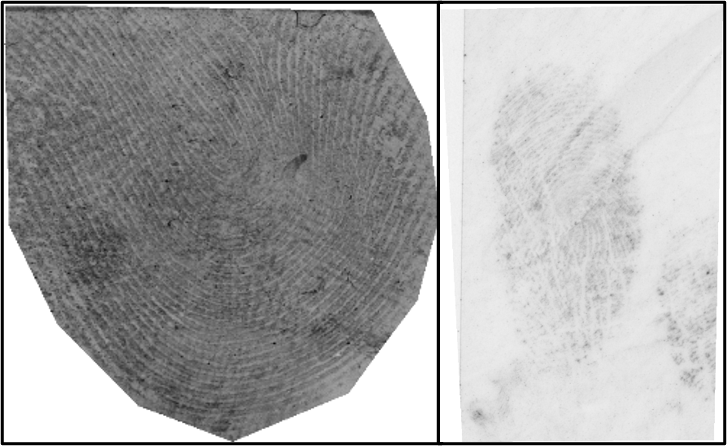}} \hfil
  \subfloat[]{\includegraphics[height=.09\linewidth]{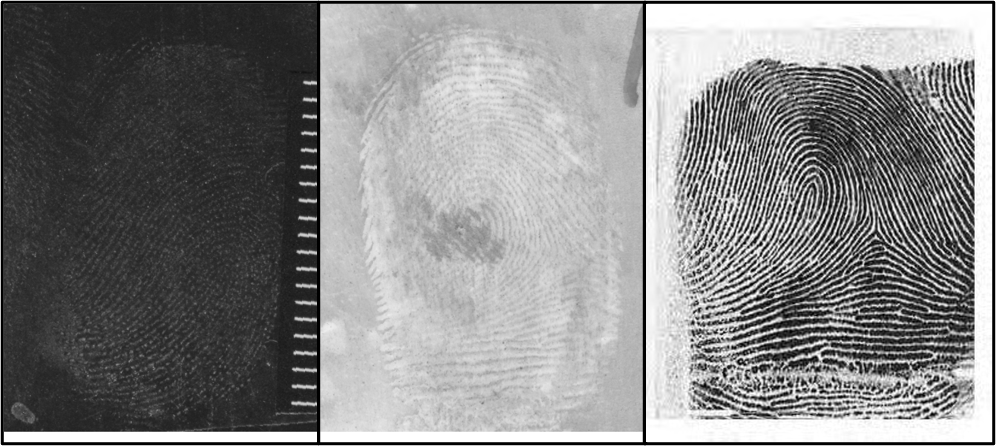}} \hfil
  \subfloat[]{\includegraphics[height=.09\linewidth]{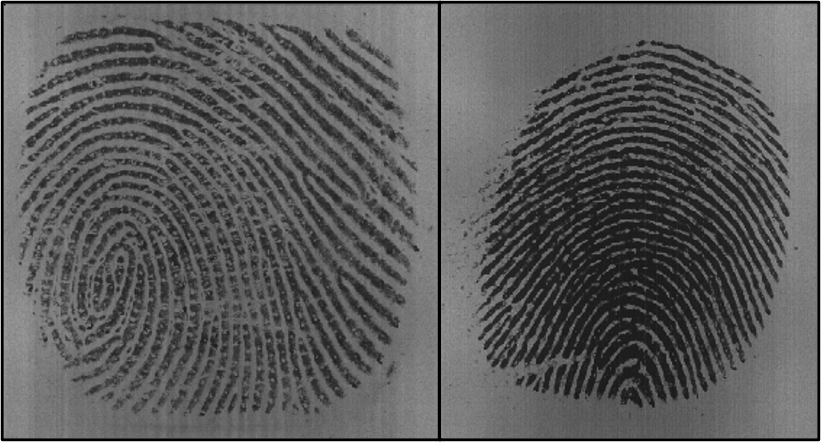}} \hfil
  \subfloat[]{\includegraphics[height=.09\linewidth]{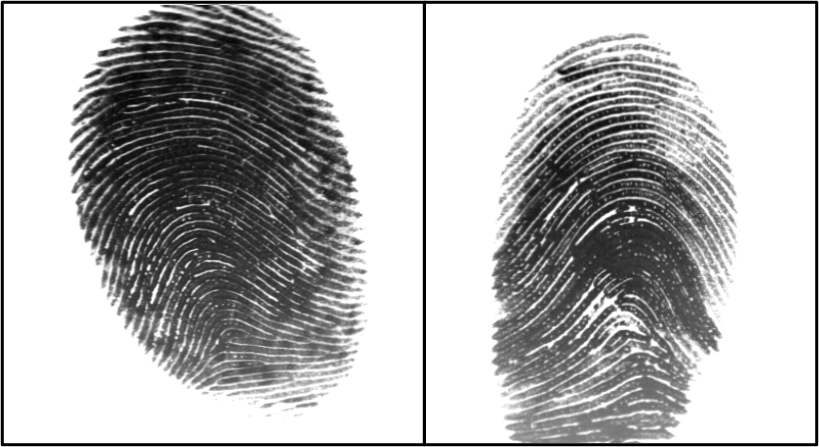}} \hfil
  \subfloat[]{\includegraphics[height=.09\linewidth]{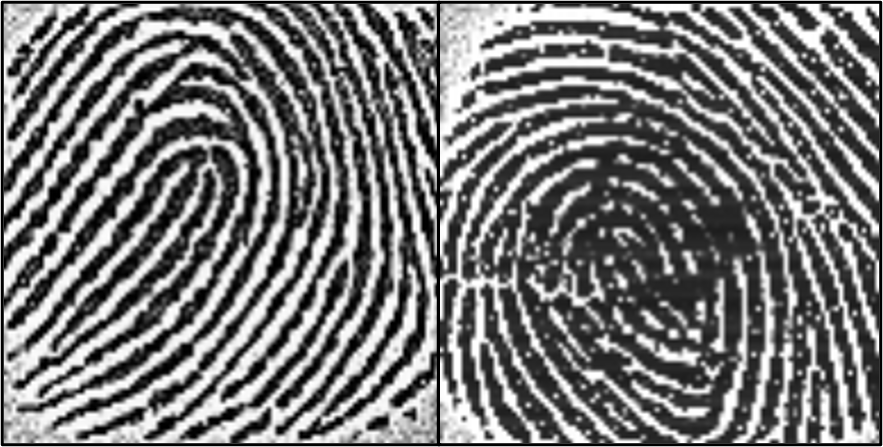}} \hfil
  \subfloat[]{\includegraphics[height=.09\linewidth]{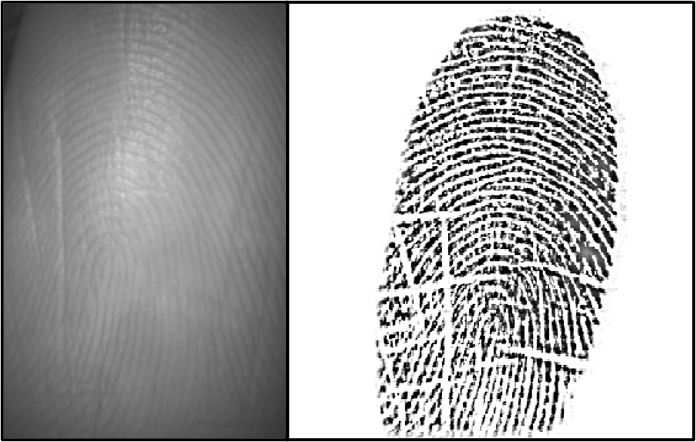}} \hfil
  \subfloat[]{\includegraphics[height=.09\linewidth]{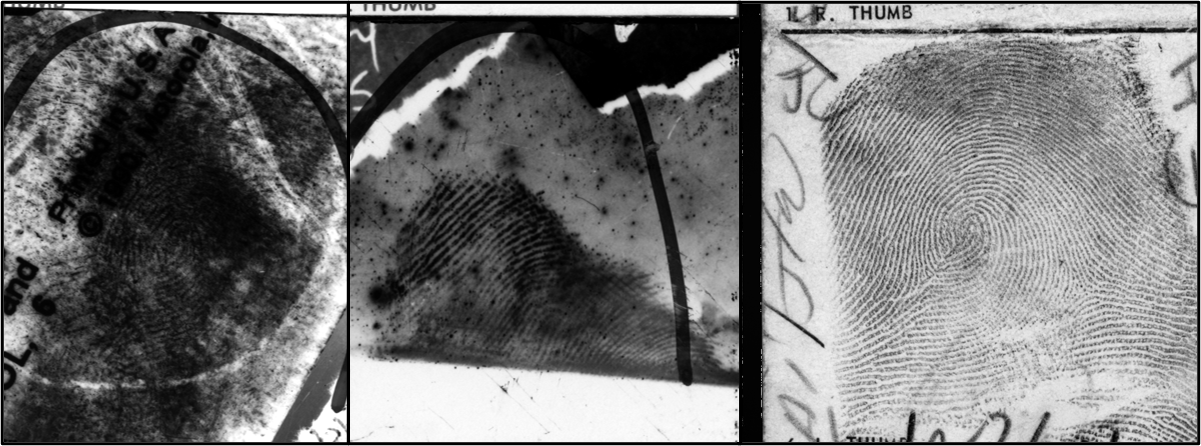}} \hfil
  \caption{Fingerprint examples from different fingerprint datasets (a) NIST SD14, (b) NIST SD4, (c) N2N Plain, (d) N2N Latent, (e) THU Latent10K (f) FVC2002 DB3A, (g) FVC2004 DB1A, (h) FVC2006 DB1A, (i) PolyU CL2CB, (j) NIST SD27. The input images are at 500 ppi, but have been rescaled in this figure to facilitate clearer visualization.} 
  \label{fig:datasets}
\end{figure*}

\section{Experiments}
\subsection{Datasets}
Aiming to validate the effectiveness and generalizability of our method, we conduct experiments on a diverse set of fingerprint datasets covering rolled, plain, partial, latent, and contactless modalities, collected under various acquisition conditions with different sensors. All fingerprint images not originally at 500 ppi are uniformly rescaled. Tab.~\ref{tab:datasets} summarizes the dataset statistics, and representative samples are illustrated in Fig.~\ref{fig:datasets}. 

We conduct training exclusively on high-quality rolled fingerprints from the NIST SD14 dataset, generating 132{,}550\footnote{Accordingly, $K$ in Sec.~III-C1 is set to 132{,}550.} minutiae-aligned patch pairs from 27{,}000 fingerprint pairs. Evaluation is then carried out directly on diverse fingerprint types without any fine-tuning. For the FVC benchmark, we select three representative datasets: FVC2002 DB3A, FVC2004 DB1A, and FVC2006 DB1A, and follow the same protocol for splitting genuine and imposter pairs as adopted in previous works~\cite{cappelli2010minutia,su2016fingerprint}. The NIST SD302 dataset~\cite{nist302} comprises 2{,}000 fingers from 200 subjects. We use subset U (2{,}000 rolled fingerprints) as the gallery and combine subsets R and S (2{,}000 plain fingerprints) as the query set, forming the N2N Plain dataset. In addition, 3{,}383 latent fingerprints are selected from subset E according to the strategy proposed by Gu et al.~\cite{gu2022latent}, and paired with subset U to form the N2N Latent dataset. For the PolyU CL2CB dataset~\cite{liu20143d}, we follow the preprocessing procedure of Cui et al.~\cite{cui2023monocular}, where contactless fingerprints are rescaled to match the mean ridge period of contact-based fingerprints.


\begin{table}[!t]
  \centering
  \vspace{-0.3cm}
  \caption{Summary of Compared Methods}
  \label{tab:compared_method}
  \renewcommand{\arraystretch}{1.06}
  \begin{threeparttable}
    \resizebox{\linewidth}{!}{
    \begin{tabular}{lll}
      \toprule
      \textbf{Method} & \textbf{Template (Anc. / Desc.)} & \textbf{Matching} \\
      \midrule
      MCC \cite{cappelli2010minutia} & V-minu.\tnote{a} / Dense & Relax. (legacy)\tnote{e} \\
      VeriFinger \cite{nist2020verifinger} & V-minu.\tnote{a} / Proprietary & Proprietary \\
      MinNet \cite{MinNet} & V-minu.\tnote{a} / 1D & Relax. (opt)\tnote{e} \\
      MSU-AFIS \cite{cao2020end} & I-minu.\tnote{b} + Ori.\tnote{d} / 1D & Graph-filter \\
      V-ExDMD \cite{pan2024latent} & V-minu.\tnote{a} / Tex. + Minu. (Dense) & Relax. (legacy)\tnote{e} \\
      V-DMD (Ours) & V-minu.\tnote{a} / Tex. + Minu. (Dense) & Relax. (opt)\tnote{e} \\
      F-DMD (Ours) & F-minu.\tnote{c} / Tex. + Minu. (Dense) & Relax. (opt)\tnote{e} \\
      \bottomrule
    \end{tabular}}
    \begin{tablenotes}
      \item[a] Minutiae extracted by VeriFinger.
      \item[b] Minutiae extracted by MSU-AFIS.
      \item[c] Minutiae extracted by FDD~\cite{FDD}.
      \item[d] Orientation field used as auxiliary anchor.
      \item[e] “Relax.” = relaxation-based matching; “opt” = optimized version.
    \end{tablenotes}
  \end{threeparttable}
\end{table}

\subsection{Compared Methods} \label{sec:compared_method}
To rigorously evaluate the proposed DMD, we benchmark it against a comprehensive set of minutiae-based fingerprint recognition methods, including MCC~\cite{cappelli2010minutia}, VeriFinger v12.0~\cite{nist2020verifinger}, MinNet~\cite{MinNet}, and MSU-AFIS~\cite{cao2020end} (Tab.~\ref{tab:compared_method}). We also evaluate three variants of DMD: our previous conference version V-ExDMD\cite{pan2024latent}, the improved V-DMD proposed in this work (both evaluated using minutiae extracted by VeriFinger), and F-DMD, which is evaluated using minutiae extracted by FDD~\cite{FDD}. All models are trained on the same training patches to ensure a fair comparison. For a fair and consistent evaluation, MCC, VeriFinger, MinNet, V-ExDMD, and V-DMD are all tested using the same set of minutiae extracted by VeriFinger v12.0. We retrain MinNet on the same set of training patches used for DMD and reimplement MCC to improve computational efficiency while maintaining its original accuracy. For the commercial matcher VeriFinger v12.0, we adopt its proprietary template format—which incorporates both minutiae and additional features—to reflect its optimal recognition performance. MSU-AFIS is evaluated as a fully integrated end-to-end system using the official code and released models, which include its native minutiae extraction, template generation, and matching procedures. 
Except for MSU-AFIS, which applies its own fingerprint enhancement as part of the end-to-end pipeline, all other methods do not incorporate any fingerprint enhancement algorithms. Tab.~\ref{tab:compared_method} summarizes all evaluated methods along with their template and matching characteristics.

\begin{table*}
  \centering
  \caption{Matching accuracy (\%) across multiple fingerprint datasets. Unless otherwise specified, TAR@FAR = 0.1\% is reported. Bold indicates the best, and italic denotes the second-best.}
  \label{tab:matching_results}
  \renewcommand{\arraystretch}{1.2}
  \begin{threeparttable}
    \resizebox{\textwidth}{!}{
      \begin{tabular}{l !{\vrule width 0.75 pt} 
        *{2}{>{\centering\arraybackslash}p{0.05\linewidth}}
        *{1}{>{\centering\arraybackslash}p{0.05\linewidth}}
        *{1}{>{\centering\arraybackslash}p{0.045\linewidth}}
        !{\vrule width 0.75 pt} 
        *{1}{>{\centering\arraybackslash}p{0.045\linewidth}}  
        *{1}{>{\centering\arraybackslash}p{0.045\linewidth}}  
        *{1}{>{\centering\arraybackslash}p{0.045\linewidth}}  
        *{1}{>{\centering\arraybackslash}p{0.045\linewidth}}  
        !{\vrule width 0.75pt} 
        *{2}{>{\centering\arraybackslash}p{0.05\linewidth}}  
        *{2}{>{\centering\arraybackslash}p{0.05\linewidth}}  
        *{2}{>{\centering\arraybackslash}p{0.05\linewidth}}  
        }
        \toprule
        \multirow{2}{*}{\textbf{Method}} 
        & \multicolumn{2}{c}{\textbf{NIST SD4}} 
        & \multicolumn{2}{c!{\vrule width 0.75 pt}}{\textbf{N2N Plain}} 
        & \textbf{FVC02}\tnote{$\ddagger$} 
        & \textbf{FVC04}\tnote{$\ddagger$} 
        & \textbf{FVC06}\tnote{$\ddagger$} 
        & \textbf{PolyU}\tnote{$\ddagger$} 
        & \multicolumn{2}{c}{\textbf{NIST SD27}} 
        & \multicolumn{2}{c}{\textbf{N2N Latent}} 
        & \multicolumn{2}{c}{\textbf{THU Latent10K}} \\
        \cmidrule(lr){2-3} \cmidrule(lr){4-5} \cmidrule(lr){6-9} \cmidrule(lr){10-11} \cmidrule(lr){12-13} \cmidrule(lr){14-15}
        & \textbf{Rank-1} & \textbf{TAR}\tnote{$\dagger$} 
        & \textbf{Rank-1} & \textbf{TAR}\tnote{$\dagger$} 
        & \textbf{TAR} & \textbf{TAR} & \textbf{TAR} & \textbf{TAR} 
        & \textbf{Rank-1} & \textbf{TAR} 
        & \textbf{Rank-1} & \textbf{TAR} 
        & \textbf{Rank-1} & \textbf{TAR} \\
        \midrule
        MCC \cite{cappelli2010minutia} & 98.60 & 98.05 & 96.20 & 92.35 & 93.11 & 85.18 & 85.23 & 39.64 & 35.27 & 13.57 & 34.94 & 19.42 & --\tnote{$\ast$} & -- \\
        VeriFinger \cite{nist2020verifinger} & 99.65 & 99.80 & 99.20 & \textbf{99.30} & \textbf{99.43} & 98.71 & 92.54 & \textbf{97.21} & 58.14 & 53.10 & 44.25 & 42.71 & -- & -- \\
        MSU-AFIS \cite{cao2020end} & 99.10 & 98.40 & 94.75 & 89.15 & 76.93 & 66.89 & 32.16 & 52.20 & 70.16 & 57.75 & 44.96 & 37.22 & 91.47 & 92.60 \\
        MinNet \cite{MinNet} & \textbf{99.85} & \textit{99.85} & \textit{99.30} & 99.20 & 98.68 & 97.61 & 93.98 & 49.21 & 67.83 & 70.93 & 47.09 & 44.69 & 87.01 & 94.70 \\
        \hhline 
        \addlinespace[0.01pt]
        \hhline 
        V-ExDMD \cite{pan2024latent} & \textit{99.80} & 99.80 & \textbf{99.40} & 99.15 & \textit{99.39} & 98.96 & 93.98 & 90.41 & 79.46 & 81.78 & 52.88 & 52.05 & 92.17 & 98.14 \\
        \rowcolor{gray!15}
        V-DMD & \textbf{99.85} & \textbf{99.90} & \textit{99.30} & 99.20 & 99.32 & 99.18 & 94.12 & 95.18 & \textit{80.62} & 81.78 & \textbf{53.21} & \textit{52.70} & 92.42 & 98.46 \\
        \rowcolor{gray!15}
        F-DMD & \textit{99.80} & 99.80 & \textit{99.30} & 99.20 & \textit{99.39} & \textit{99.57} & \textit{95.00} & \textit{95.23} & \textbf{82.56} & \textbf{85.27} & \textit{53.15} & \textbf{52.79} & \textbf{92.80} & \textbf{98.54} \\
        \rowcolor{gray!15}
        F-DMD-B\tnote{$\S$} & \textit{99.80} & 99.80 & 99.25 & \textit{99.25} & 99.11 & \textbf{99.71} & \textbf{95.10} & 95.01 & 79.07 & \textit{83.72} & 52.53 & 51.82 & \textit{92.57} & \textit{98.49} \\
        \bottomrule
      \end{tabular}
    }
    \begin{tablenotes}
      \item[$\dagger$] TAR@FAR = 0.01\%.
      \item[$\ddagger$] FVC02, FVC04, FVC06, and PolyU refer to FVC2002 DB3A, FVC2004 DB1A, FVC2006 DB1A, and PolyU CL2CB, respectively.
      \item[$\ast$] ``--'' indicates results not available due to excessive runtime or computational cost.
      \item[$\S$] F-DMD-B is a binarized variant of F-DMD, where descriptor and mask values are thresholded.
    \end{tablenotes}
  \end{threeparttable}
\end{table*}

\begin{figure*}[!t]
  \centering
  \subfloat[NIST SD27]{\includegraphics[height=.24\linewidth]{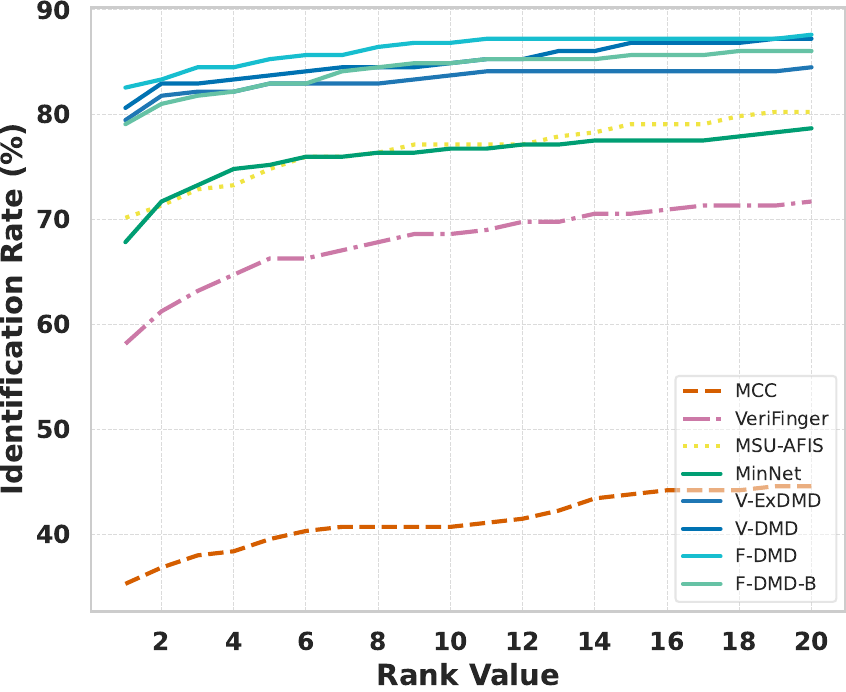}} \hfil
  \subfloat[N2N Latent]{\includegraphics[height=.24\linewidth]{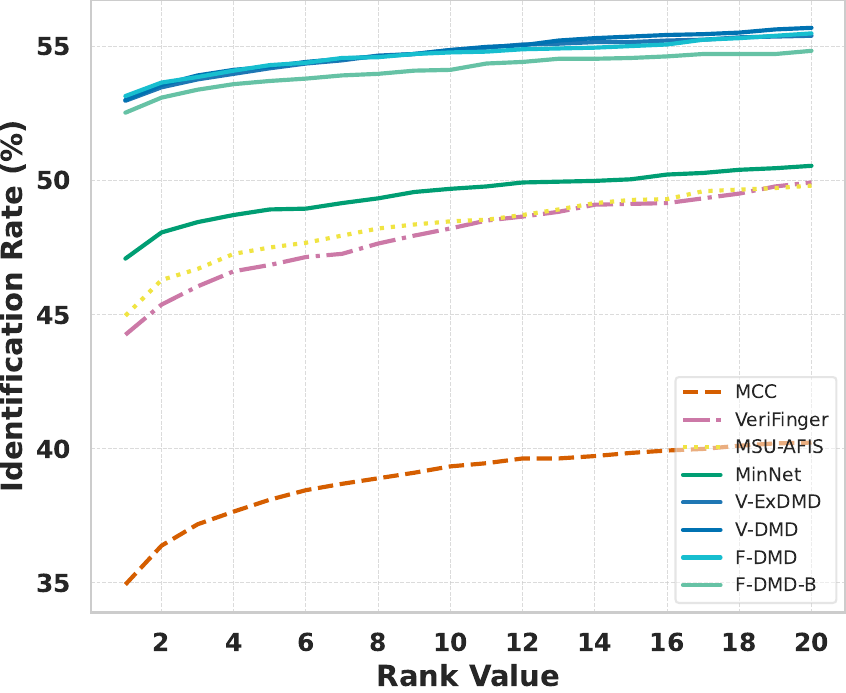}} \hfil
  \subfloat[THU Latent10K]{\includegraphics[height=.24\linewidth]{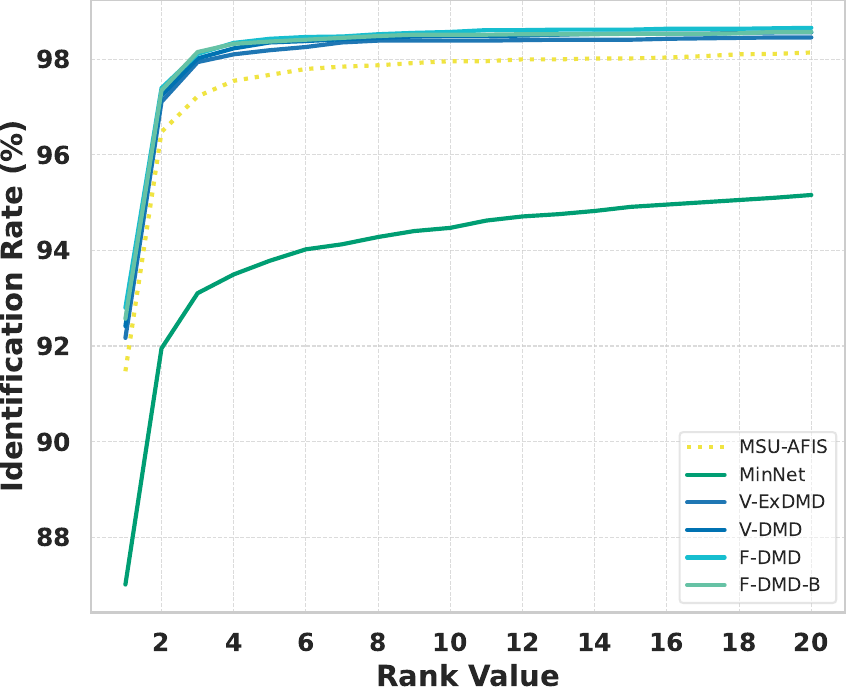}} \hfil
  \caption{CMC curves for latent fingerprint matching on (a) NIST SD27, (b) N2N Latent, and (c) THU Latent10K, respectively.}
  \label{fig:cmc_curves}
\end{figure*}

\begin{figure*}[!t]
  \centering
  \subfloat[NIST SD27]{\includegraphics[height=.24\linewidth]{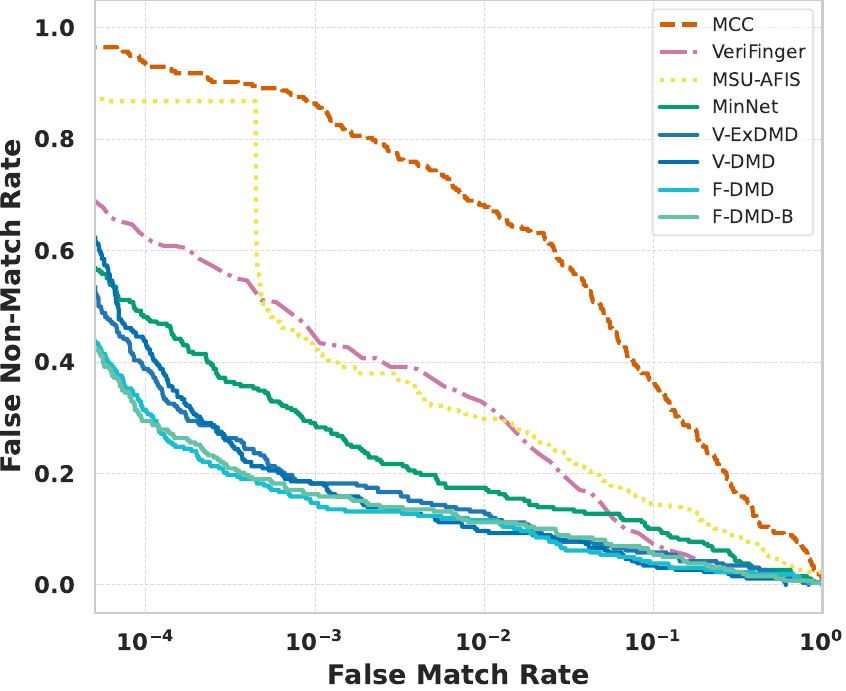}} \hfil
  \subfloat[N2N Latent]{\includegraphics[height=.24\linewidth]{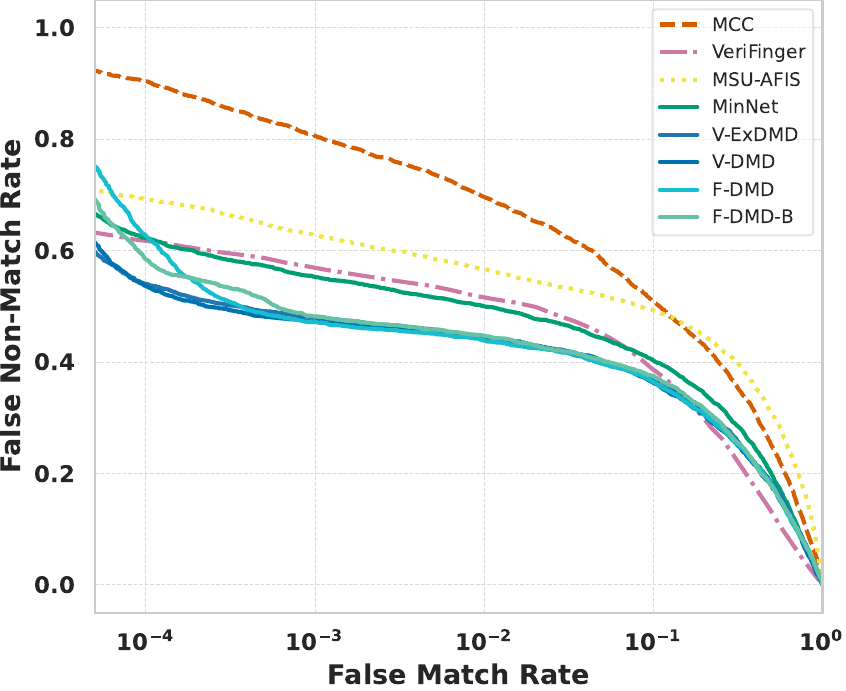}} \hfil
  \subfloat[THU Latent10K]{\includegraphics[height=.24\linewidth]{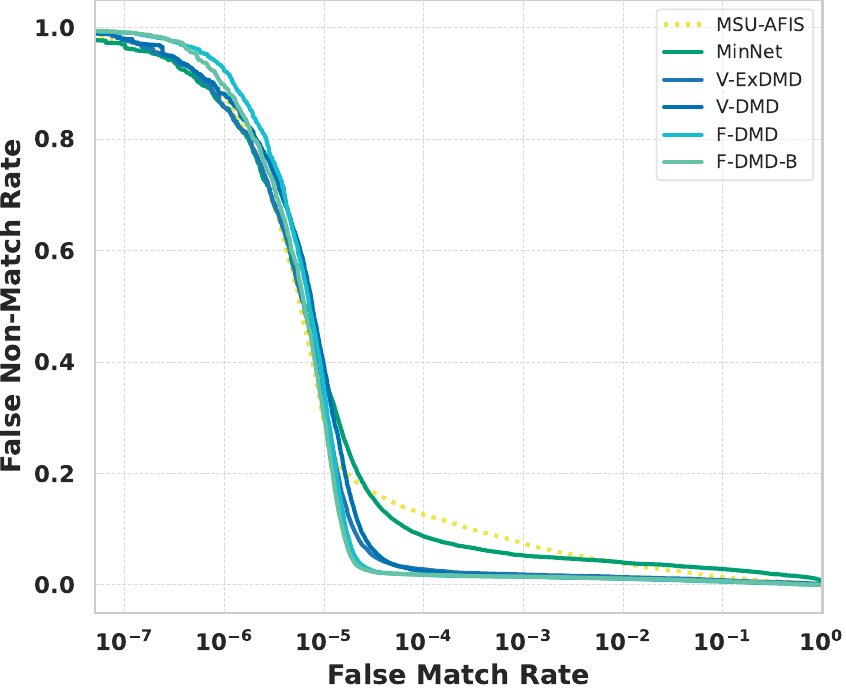}} \hfil
  \caption{DET curves for latent fingerprint matching on (a) NIST SD27, (b) N2N Latent, and (c) THU Latent10K, respectively.}
  \label{fig:det_curves}
\end{figure*}

\subsection{Fingerprint Matching Performance} \label{sec:match_eval}
The proposed DMD is benchmarked against a series of representative methods, as summarized in Tab.~\ref{tab:compared_method}. Rank-1 accuracy and TAR@FAR=0.1\%(0.01\%) are adopted as the primary metrics for closed-set and open-set evaluations, respectively. In addition, Cumulative Match Characteristic (CMC) curves and Detection Error Tradeoff (DET) curves are included to provide comprehensive quantitative comparisons. 
For V-ExDMD~\cite{pan2024latent} and V-DMD, which both rely on VeriFinger-extracted minutiae, we apply the score standardization strategy from~\cite{pan2024latent} to mitigate the impact of spurious minutiae near the fingerprint boundary. In contrast, F-DMD does not use this normalization, as FDD tends to produce minutiae well-confined within the fingerprint foreground. Moreover, we further extend F-DMD by introducing a binary variant. In this setting, the local dense representation $f$ is binarized using a threshold of 0, and the foreground segmentation map $h$ is binarized at 0.5. 

\begin{figure*}[!t]
  \centering
  \includegraphics[width=.95\linewidth]{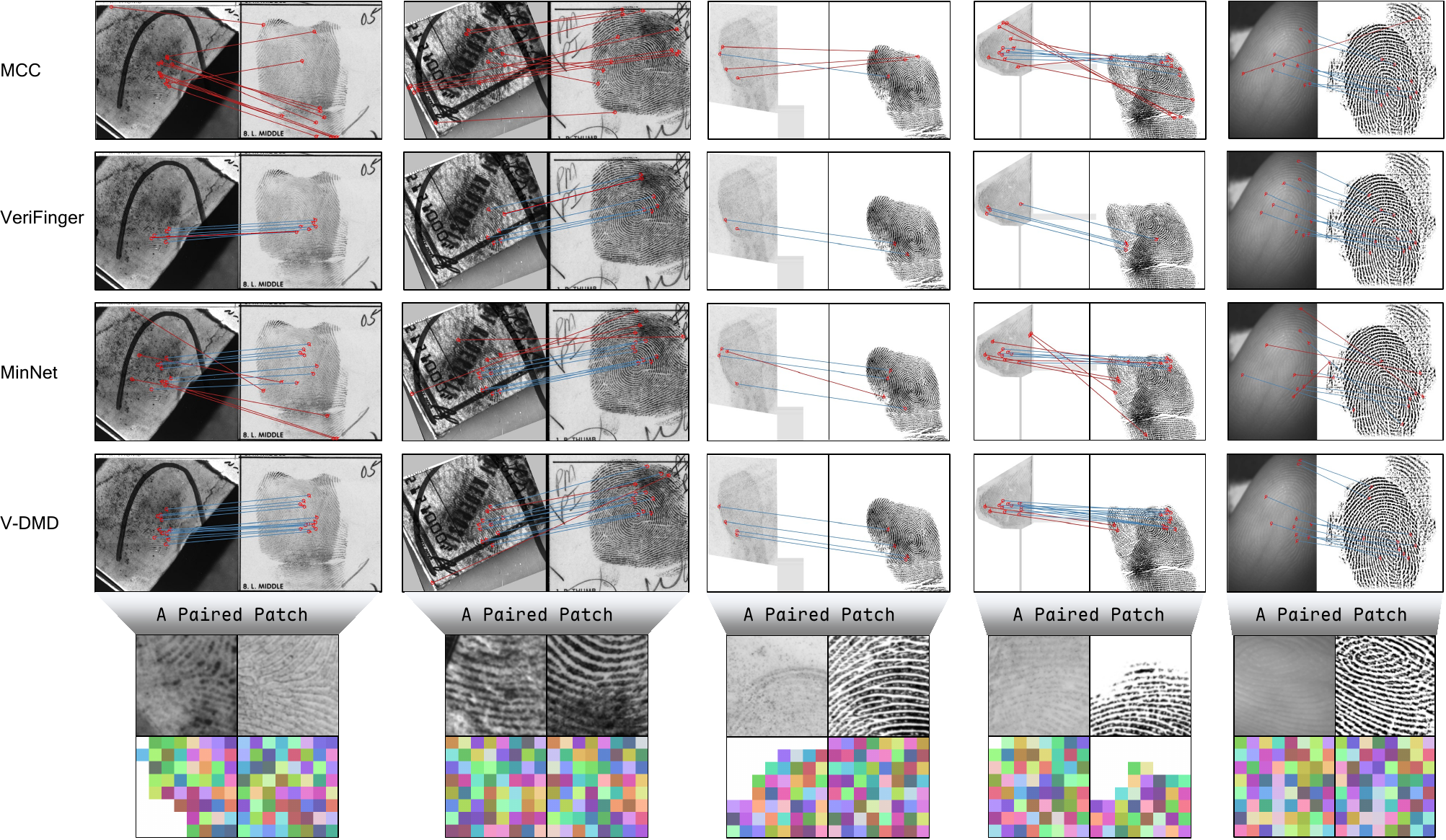}
  \caption{Comparison of minutiae matching performance and V-DMD feature visualization. The top part shows the matched minutiae pairs used for similarity computation, the correct matches are colored blue line and the incorrect matches are colored red. For each example, one matched patch pair is selected, and their corresponding DMD features are visualized below, illustrating the effectiveness of the local dense representation.}
  \label{fig:mnt_matching}
  \vspace{-0.3cm}
\end{figure*}

Tab.~\ref{tab:matching_results} summarizes the matching performance across a range of fingerprint datasets, arranged approximately from high-quality rolled/plain fingerprints to more challenging cases, inculding partial, distorted, cross-modality, and latent fingerprints. Conventional methods such as MCC\cite{cappelli2010minutia}, which rely on hand-crafted features, are highly sensitive to noise and partial overlaps that degrade minutiae quality, resulting in low robustness. The commercial matcher VeriFinger~\cite{nist2020verifinger} performs well on high-quality inputs but struggles with low-quality or latent fingerprints. Furthermore, both MCC and VeriFinger exhibit limited scalability due to slow matching speeds, making them impractical for large-scale retrieval (see Sec.\ref{sec:effi}). MSU-AFIS\cite{cao2020end}, developed specifically for latent fingerprint matching, achieves reasonable performance on forensic data but generalizes poorly to other modalities. MinNet~\cite{MinNet} shows more balanced performance overall, yet underperforms in cross-modality scenarios such as contactless-to-rolled matching.

In contrast, our DMD-based methods demonstrate consistently strong performance across all datasets, thanks to their local dense representation that jointly encodes texture and minutiae-related features. Compared to V-ExDMD~\cite{pan2024latent}, V-DMD incorporates  patches standardization layer and low-quality data augmentations, enhancing performance on latent and contactless fingerprints while maintaining high accuracy elsewhere. As to F-DMD, which replaces VeriFinger minutiae with those extracted using FDD~\cite{FDD}, this variant improves robustness in challenging cases due to better foreground minutiae detection. Notably, the binarized version F-DMD-B achieves similar accuracy while significantly reducing storage and boosting speed (see Sec.\ref{sec:effi}), making it suitable for large-scale fingerprint identification.

To complement the quantitative results in Tab.~\ref{tab:matching_results}, we further visualize the performance on the three latent fingerprint datasets using CMC and DET curves in Figs.~\ref{fig:cmc_curves} and~\ref{fig:det_curves}. These curves offer a more detailed comparison under realistic forensic conditions and consistently demonstrate the superiority of our DMD-based methods in terms of both identification accuracy and verification robustness. 

Furthermore, Fig.\ref{fig:mnt_matching} presents several examples of minutiae-aligned patch matching across different local representation methods. Since MSU-AFIS\cite{cao2020end} utilizes its own minutiae and orientation field extraction pipeline to define local anchors, its results are not included in the visualization. All other comparisons are based on minutiae extracted using VeriFinger v12.0~\cite{nist2020verifinger} for consistency. As shown, compared methods tend to produce excessive false matches \cite{cappelli2010minutia,MinNet} or adopt overly conservative strategies \cite{nist2020verifinger}. In contrast, DMD achieves both high match accuracy and broad coverage by involving a greater number of minutiae in the score computation. This advantage stems from its use of local dense representations. The bottom part of Fig.~\ref{fig:mnt_matching} further visualizes the DMD features for selected matched patch pairs. The features are activated only within valid foreground regions, highlighting DMD's spatial awareness. Even under challenging conditions such as noisy latent prints or modality discrepancies in contactless fingerprints, DMD exhibits strong consistency in its feature representations, demonstrating robustness and generalization across domains. 

\subsection{Ablation Study}
To further validate the effectiveness of the proposed DMD design, we conduct ablation studies on three degenerated variants: 1. DMD w/o Minutiae Stream (DMD$^{-M}$): This variant removes the Minutiae Stream entirely. 2. Single-Stream DMD (DMD$^{S}$): This variant merges the minutiae and texture streams into a single unified stream. 3. DMD w/o Positional Embedding (DMD$^{-P}$): This variant removes the position embedding introduced in the dense representation extraction stage. For a fair comparison, all variants use the same local dense representation dimension, with $f \in \mathbb{R}^{12 \times 8 \times 8}$. Evaluation is conducted using minutiae extracted by VeriFinger, consistent with the experimental setting used for V-DMD.

\begin{table}[!t]
  \centering
  \caption{Ablation evaluations on the extraction of DMD. TAR@FAR=0.1\% is reported.}
  \label{tab:ablation}
  \begin{tabular}{l
    *{1}{>{\centering\arraybackslash}p{0.12\linewidth}}
    *{1}{>{\centering\arraybackslash}p{0.06\linewidth}}
    *{1}{>{\centering\arraybackslash}p{0.12\linewidth}}
    *{1}{>{\centering\arraybackslash}p{0.06\linewidth}}
    *{1}{>{\centering\arraybackslash}p{0.12\linewidth}}
    *{1}{>{\centering\arraybackslash}p{0.06\linewidth}}
  }
  \toprule
  \multirow{2}{*}{\textbf{Ablations}} & \multicolumn{2}{c}{\textbf{NIST SD27}} & \multicolumn{2}{c}{\textbf{N2N Latent}} & \multicolumn{2}{c}{\textbf{THU Latent10K}} \\
  \cmidrule(lr){2-3} \cmidrule(lr){4-5} \cmidrule(lr){6-7}
  & \textbf{Rank-1} & \textbf{TAR} & \textbf{Rank-1} & \textbf{TAR} & \textbf{Rank-1} & \textbf{TAR} \\
  \midrule
  DMD$^{-M}$ & 67.44 & 70.93 & 51.70 & 50.52 & 78.09 & 86.09 \\
  DMD$^{S}$ & 72.87 & 78.29 & 52.02 & 51.40 & 85.07 & 92.36 \\
  DMD$^{-P}$ & 79.84 & \textbf{82.56} & 52.94 & 52.35 & 92.36 & 98.44 \\
  \hhline
  DMD & \textbf{80.62} & 81.78 & \textbf{53.21} & \textbf{52.70} & \textbf{92.42} & \textbf{98.46} \\
  \bottomrule
  \end{tabular}
   \vspace{-0.3cm}
\end{table}

Tab.~\ref{tab:ablation} summarizes the results of the ablation study evaluating the impact of key components in DMD. Removing the Minutiae Stream (DMD$^{-M}$) leads to the most significant performance drop across all datasets, indicating the crucial role of minutiae-guided features in fingerprint matching. Merging the two streams into a unified representation (DMD$^{S}$) also causes a noticeable decline, demonstrating that modeling texture and minutiae cues in separate branches leads to more discriminative features. Eliminating the position embedding (DMD$^{-P}$) results in slight fluctuations across datasets; however, its overall contribution remains positive, as it helps encode spatial context that is beneficial for robust matching under challenging conditions. Overall, the full DMD configuration consistently outperforms its ablated variants, validating the effectiveness of the proposed design.

\subsection{Efficiency} \label{sec:effi}

Tab.~\ref{tab:effi} compares the efficiency of different methods in terms of model size, template compactness, feature extraction speed, and matching throughput, all measured on the NIST SD27 dataset. The extraction time includes minutiae detection, descriptor extraction, and template serialization. MCC \cite{cappelli2010minutia}, VeriFinger \cite{nist2020verifinger}, MinNet \cite{MinNet}, and V-DMD all rely on minutiae extracted by VeriFinger v12.0 in our experimental setting, which involves a relatively slow detection process and therefore results in consistently limited extraction speed (~0.45 fps). MSU-AFIS, despite employing quantization, shows even lower efficiency due to the additional estimation of dense orientation-based anchors and corresponding descriptor generation, leading to larger templates and slower processing.

Motivated by the need for higher efficiency, we evaluate two variants—F-DMD and its binarized counterpart F-DMD-B. By leveraging the foreground-aware minutiae extractor from FDD\cite{FDD}, these methods significantly reduce the number of detected minutiae, resulting in smaller templates and faster matching speed. Notably, F-DMD-B further compresses the template to an average of just 4.45 KB and achieves the highest matching throughput (5,305 pairs/s) among all compared methods. Importantly, these efficiency gains come without compromising accuracy. As shown in Tab.~\ref{tab:matching_results}, F-DMD-B maintains strong matching performance, making it particularly well-suited for large-scale identification scenarios where speed and storage are critical.

\begin{table}[!t]
  \centering
  \caption{Comparison of model size, template compactness, and runtime efficiency across methods.}
  \label{tab:effi}
  \begin{threeparttable}
    \resizebox{\linewidth}{!}{
  \begin{tabular}{l
    *{1}{>{\centering\arraybackslash}p{0.15\linewidth}}
    *{1}{>{\centering\arraybackslash}p{0.15\linewidth}}
    *{1}{>{\centering\arraybackslash}p{0.15\linewidth}}
    *{1}{>{\centering\arraybackslash}p{0.15\linewidth}}
    }
    \toprule
    \textbf{Method}\tnote{$\dagger$} & 
    \makecell[c]{\textbf{\#Params} \\ (M)} & 
    \makecell[c]{\textbf{Template} \\ (KB)} & 
    \makecell[c]{\textbf{Extraction} \\ (fps)} & 
    \makecell[c]{\textbf{Matching} \\ (pairs/s)}\\
    \midrule
    MCC \cite{cappelli2010minutia} & --- & 255.39 & 0.43 & 212.12 \\
    VeriFinger \cite{nist2020verifinger} & --- & 3.37 & 0.45 & 52.76 \\
    MSU-AFIS \cite{cao2020end} & 14.70 & 516.92 & 0.05 & 1,413.91 \\
    MinNet & 6.57 & 144.12 & 0.45 & 2,564.10 \\
    \hhline
    V-DMD & 43.16 & 150.11 & 0.45 & 2,469.14 \\
    F-DMD & 43.16 & 129.60 & 4.17 & 4,814.64 \\
    F-DMD-B & 43.16 & 4.45 & 4.17 & 5,305.04 \\
    \bottomrule
  \end{tabular}}
  \begin{tablenotes}
    \item[$\dagger$] Average template size, extraction speed, and matching throughput are \\ measured on the NIST SD27 dataset for all methods.
  \end{tablenotes}
  \end{threeparttable}
  \vspace{-0.3cm}
\end{table}

\section{Discussion}
Our approach builds upon a multi-level, fine-grained dense representation DMD to enhance the spatial expressiveness of local descriptors. In addition, it integrates the FDD framework \cite{FDD} for efficient minutiae extraction and reimplements the relaxation-based score refinement module to improve computational efficiency. These design choices aim to improve both the descriptive capacity and practical performance of the system, contributing to better accuracy and faster matching in a variety of fingerprint recognition scenarios.

Nonetheless, there remains room for improvement in the matching module. Some false matches arise from clusters of minutiae being incorrectly aligned to small local fingerprint regions. Future work will explore incorporating both the number and spatial distribution of matched minutiae into the score refinement process. In addition, we aim to investigate the consistency between minutiae-level correspondences and global fingerprint pose estimation, which may help further reduce erroneous matches. We also plan to integrate fixed-length and minutiae-based dense descriptors to fully leverage their complementary strengths, with the goal of developing a more accurate and efficient fingerprint recognition system based on dense representations. Finally, as the current DMD extraction remains CNN-based, we plan to explore Transformer architectures to enhance representation robustness, inspired by the recent success of large vision-language models \cite{CLIP}. 

\section{Conclusion}
This paper presents DMD, a minutiae-anchored local descriptor that introduces a spatially structured dense representation for robust fingerprint matching. In contrast to traditional representations that rely on flattened vectors, DMD preserves the spatial arrangement of local features within minutia-centered patches, enabling multi-level, fine-grained representation of both ridge texture and minutiae context. In addition, the spatial structure allows DMD to focus on valid foreground regions within each patch, effectively suppressing background noise and improving matching quality. A dual-stream architecture captures complementary features, while a lightweight relaxation-based refinement enhances structural consistency in the matching stage. To ensure practical efficiency, we incorporate an efficient foreground-aware minutiae extraction pipeline and apply binarization to reduce template size without compromising accuracy. Extensive experiments demonstrate that DMD achieves state-of-the-art performance across rolled, contactless, and latent fingerprints, while maintaining high computational efficiency—highlighting its potential for real-world deployment.

\bibliographystyle{IEEEtran}
\bibliography{IEEEabrv,refs} %

\begin{thebibliography}{10}
\providecommand{\url}[1]{#1}
\csname url@samestyle\endcsname
\providecommand{\newblock}{\relax}
\providecommand{\bibinfo}[2]{#2}
\providecommand{\BIBentrySTDinterwordspacing}{\spaceskip=0pt\relax}
\providecommand{\BIBentryALTinterwordstretchfactor}{4}
\providecommand{\BIBentryALTinterwordspacing}{\spaceskip=\fontdimen2\font plus
\BIBentryALTinterwordstretchfactor\fontdimen3\font minus
  \fontdimen4\font\relax}
\providecommand{\BIBforeignlanguage}[2]{{%
\expandafter\ifx\csname l@#1\endcsname\relax
\typeout{** WARNING: IEEEtran.bst: No hyphenation pattern has been}%
\typeout{** loaded for the language `#1'. Using the pattern for}%
\typeout{** the default language instead.}%
\else
\language=\csname l@#1\endcsname
\fi
#2}}
\providecommand{\BIBdecl}{\relax}
\BIBdecl

\bibitem{maltoni2022handbook}
D.~Maltoni, D.~Maio, A.~K. Jain, and J.~Feng, \emph{Handbook of Fingerprint
  Recognition}, 3rd~ed.\hskip 1em plus 0.5em minus 0.4em\relax Springer Nature,
  Cham, Switzerland, 2022.

\bibitem{cole2004history}
S.~A. Cole, ``History of fingerprint pattern recognition,'' in \emph{Automatic
  fingerprint recognition systems}.\hskip 1em plus 0.5em minus 0.4em\relax
  Springer, 2004, pp. 1--25.

\bibitem{munoz2013fingerprint}
A.~Munoz-Briseno, A.~Gago-Alonso, and J.~Hern{\'a}ndez-Palancar, ``Fingerprint
  indexing with bad quality areas,'' \emph{Expert Systems with Applications},
  vol.~40, no.~5, pp. 1839--1846, 2013.

\bibitem{iloanusi2011indexing}
O.~Iloanusi, A.~Gyaourova, and A.~Ross, ``Indexing fingerprints using minutiae
  quadruplets,'' in \emph{CVPR 2011 WORKSHOPS}.\hskip 1em plus 0.5em minus
  0.4em\relax IEEE, 2011, pp. 127--133.

\bibitem{cappelli2010minutia}
R.~Cappelli, M.~Ferrara, and D.~Maltoni, ``{Minutia Cylinder-Code}: A new
  representation and matching technique for fingerprint recognition,''
  \emph{IEEE Transactions on Pattern Analysis and Machine Intelligence},
  vol.~32, no.~12, pp. 2128--2141, 2010.

\bibitem{Tico2003fingerprint}
M.~Tico and P.~Kuosmanen, ``Fingerprint matching using an orientation-based
  minutia descriptor,'' \emph{IEEE Transactions on Pattern Analysis and Machine
  Intelligence}, vol.~25, no.~8, pp. 1009--1014, 2003.

\bibitem{feng2008combining}
J.~Feng, ``Combining minutiae descriptors for fingerprint matching,''
  \emph{Pattern Recognition}, vol.~41, no.~1, pp. 342--352, 2008.

\bibitem{zhou2013fingerprint}
R.~Zhou, D.~Zhong, and J.~Han, ``Fingerprint identification using sift-based
  minutia descriptors and improved all descriptor-pair matching,''
  \emph{Sensors}, vol.~13, no.~3, pp. 3142--3156, 2013.

\bibitem{cao2018texture}
K.~Cao and A.~K. Jain, ``Latent fingerprint recognition: Role of texture
  template,'' in \emph{2018 IEEE 9th International Conference on Biometrics
  Theory, Applications and Systems (BTAS)}, 2018, pp. 1--9.

\bibitem{cao2019automated}
------, ``Automated latent fingerprint recognition,'' \emph{IEEE Transactions
  on Pattern Analysis and Machine Intelligence}, vol.~41, no.~4, pp. 788--800,
  2019.

\bibitem{cao2020end}
K.~Cao, D.-L. Nguyen, C.~Tymoszek, and A.~K. Jain, ``End-to-end latent
  fingerprint search,'' \emph{IEEE Transactions on Information Forensics and
  Security}, vol.~15, pp. 880--894, 2020.

\bibitem{MinNet}
H.~{\.I}. {\"O}zt{\"u}rk, B.~Selbes, and Y.~Artan, ``{MinNet}: {M}inutia patch
  embedding network for automated latent fingerprint recognition,'' in
  \emph{2022 IEEE/CVF Conference on Computer Vision and Pattern Recognition
  Workshops}, 2022, pp. 1626--1634.

\bibitem{grosz2023latent}
S.~A. Grosz and A.~K. Jain, ``Latent fingerprint recognition: {F}usion of local
  and global embeddings,'' \emph{IEEE Transactions on Information Forensics and
  Security}, vol.~18, pp. 5691--5705, 2023.

\bibitem{DeepPrint}
J.~J. Engelsma, K.~Cao, and A.~K. Jain, ``Learning a fixed-length fingerprint
  representation,'' \emph{IEEE Transactions on Pattern Analysis and Machine
  Intelligence}, vol.~43, no.~6, pp. 1981--1997, 2021.

\bibitem{nist2020verifinger}
\BIBentryALTinterwordspacing
{Neurotechnology}. {VeriFinger SDK}. Accessed: Apr. 18, 2025. [Online].
  Available: \url{https://www.neurotechnology.com/verifinger.html}
\BIBentrySTDinterwordspacing

\bibitem{pan2024latent}
Z.~Pan, Y.~Duan, X.~Guan, J.~Feng, and J.~Zhou, ``Latent fingerprint matching
  via dense minutia descriptor,'' in \emph{2024 IEEE International Joint
  Conference on Biometrics (IJCB)}, 2024, pp. 1--10.

\bibitem{grosz2024afrnet}
S.~A. Grosz and A.~K. Jain, ``{AFR-Net}: Attention-driven fingerprint
  recognition network,'' \emph{IEEE Transactions on Biometrics, Behavior, and
  Identity Science}, vol.~6, no.~1, pp. 30--42, 2024.

\bibitem{gu2022latent}
S.~Gu, J.~Feng, J.~Lu, and J.~Zhou, ``Latent fingerprint indexing: Robust
  representation and adaptive candidate list,'' \emph{IEEE Transactions on
  Information Forensics and Security}, vol.~17, pp. 908--923, 2022.

\bibitem{FDD}
Z.~Pan, Y.~Duan, J.~Feng, and J.~Zhou, ``Fixed-length dense descriptor for
  efficient fingerprint matching,'' in \emph{2024 IEEE International Workshop
  on Information Forensics and Security (WIFS)}, 2024, pp. 1--6.

\bibitem{PFVNet}
Z.~He, J.~Zhang, L.~Pang, and E.~Liu, ``{PFVNet}: {A} partial fingerprint
  verification network learned from large fingerprint matching,'' \emph{IEEE
  Transactions on Information Forensics and Security}, vol.~17, pp. 3706--3719,
  2022PFV.

\bibitem{guan2025joint}
X.~Guan, Z.~Pan, J.~Feng, and J.~Zhou, ``Joint identity verification and pose
  alignment for partial fingerprints,'' \emph{IEEE Transactions on Information
  Forensics and Security}, vol.~20, pp. 249--263, 2025.

\bibitem{ifvit}
Y.~Qiu, H.~Chen, X.~Dong, Z.~Lin, I.~Yi~Liao, M.~Tistarelli, and Z.~Jin,
  ``{IFViT}: Interpretable fixed-length representation for fingerprint matching
  via vision transformer,'' \emph{IEEE Transactions on Information Forensics
  and Security}, vol.~20, pp. 559--573, 2025.

\bibitem{gu2021latent}
S.~Gu, J.~Feng, J.~Lu, and J.~Zhou, ``Latent fingerprint registration via
  matching densely sampled points,'' \emph{IEEE Transactions on Information
  Forensics and Security}, vol.~16, pp. 1231--1244, 2021.

\bibitem{resnet}
K.~He, X.~Zhang, S.~Ren, and J.~Sun, ``Deep residual learning for image
  recognition,'' in \emph{Proceedings of the IEEE conference on computer vision
  and pattern recognition}, 2016, pp. 770--778.

\bibitem{vaswani2017attention}
A.~Vaswani, N.~Shazeer, N.~Parmar, J.~Uszkoreit, L.~Jones, A.~N. Gomez,
  {\L}.~Kaiser, and I.~Polosukhin, ``Attention is all you need,''
  \emph{Advances in Neural Information Processing Systems}, vol.~30, 2017.

\bibitem{wang2018cosface}
H.~Wang, Y.~Wang, Z.~Zhou, X.~Ji, D.~Gong, J.~Zhou, Z.~Li, and W.~Liu,
  ``Cosface: Large margin cosine loss for deep face recognition,'' in
  \emph{Proceedings of the IEEE conference on computer vision and pattern
  recognition}, 2018, pp. 5265--5274.

\bibitem{duan2023estimating}
Y.~Duan, J.~Feng, J.~Lu, and J.~Zhou, ``Estimating fingerprint pose via dense
  voting,'' \emph{IEEE Transactions on Information Forensics and Security},
  vol.~18, pp. 2493--2507, 2023.

\bibitem{si2015detection}
X.~Si, J.~Feng, J.~Zhou, and Y.~Luo, ``Detection and rectification of distorted
  fingerprints,'' \emph{IEEE Transactions on Pattern Analysis and Machine
  Intelligence}, vol.~37, no.~3, pp. 555--568, 2015.

\bibitem{FLARE}
Z.~Pan, X.~Guan, Y.~Duan, J.~Feng, and J.~Zhou, ``Fixed-length dense
  fingerprint representation,'' \emph{arXiv preprint arXiv:2505.03597}, 2025.

\bibitem{su2016fingerprint}
Y.~Su, J.~Feng, and J.~Zhou, ``Fingerprint indexing with pose constraint,''
  \emph{Pattern Recognition}, vol.~54, pp. 1--13, 2016.

\bibitem{nist302}
G.~Fiumara, P.~Flanagan, J.~Grantham, K.~Ko, K.~Marshall, M.~Schwarz,
  E.~Tabassi, B.~Woodgate, and C.~Boehnen, ``National institute of standards
  and technology special database 302: Nail to nail fingerprint challenge,''
  \emph{Technical Note 2007, National Institute of Standards and Technology},
  2018.

\bibitem{liu20143d}
F.~Liu and D.~Zhang, ``{3D} fingerprint reconstruction system using feature
  correspondences and prior estimated finger model,'' \emph{Pattern
  Recognition}, vol.~47, no.~1, pp. 178--193, 2014.

\bibitem{cui2023monocular}
Z.~Cui, J.~Feng, and J.~Zhou, ``Monocular {3D} fingerprint reconstruction and
  unwarping,'' \emph{IEEE Transactions on Pattern Analysis and Machine
  Intelligence}, vol.~45, no.~7, pp. 8679--8695, 2023.

\bibitem{CLIP}
A.~Radford, J.~W. Kim, C.~Hallacy, A.~Ramesh, G.~Goh, S.~Agarwal, G.~Sastry,
  A.~Askell, P.~Mishkin, J.~Clark \emph{et~al.}, ``Learning transferable visual
  models from natural language supervision,'' in \emph{International conference
  on machine learning}.\hskip 1em plus 0.5em minus 0.4em\relax PmLR, 2021, pp.
  8748--8763.

\end{thebibliography}

\end{document}